\documentclass{article}
\usepackage{spconf,amsmath,graphicx}
\usepackage{multirow}
\usepackage{booktabs}
\usepackage[hyperfootnotes=false]{hyperref}

\title{Tackling data scarcity in Speech Translation using zero-shot multilingual Machine Translation techniques}
\name{Tu Anh Dinh, Danni Liu, Jan Niehues\thanks{This paper was done in collaboration with Mediaan, Heerlen.\newline Copyright 2022 IEEE. Published in ICASSP 2022 - 2022 IEEE International Conference on Acoustics, Speech and Signal Processing (ICASSP), scheduled for 22-27 May 2022 in Singapore. Personal use of this material is permitted. However, permission to reprint/republish this material for advertising or promotional purposes or for creating new collective works for resale or redistribution to servers or lists, or to reuse any copyrighted component of this work in other works, must be obtained from the IEEE. Contact: Manager, Copyrights and Permissions / IEEE Service Center / 445 Hoes Lane / P.O. Box 1331 / Piscataway, NJ 08855-1331, USA. Telephone: + Intl. 908-562-3966.}}
\address{Department of Data Science and Knowledge Engineering\\ Maastricht University, The Netherlands}
\begin{document}
%
\maketitle
\begin{abstract}
		Recently, end-to-end speech translation (ST) has gained significant attention as it avoids error propagation. However, the approach suffers from data scarcity. It heavily depends on direct ST data and is less efficient in making use of speech transcription and text translation data, which is often more easily available. In the related field of multilingual text translation, several techniques have been proposed for zero-shot translation. A main idea is to increase the similarity of semantically similar sentences in different languages. We investigate whether these ideas can be applied to speech translation, by building ST models trained on speech transcription and text translation data. We investigate the effects of data augmentation and auxiliary loss function. The techniques were successfully applied to few-shot ST using limited ST data, with improvements of up to +12.9 BLEU points compared to direct end-to-end ST and +3.1 BLEU points compared to ST models fine-tuned from ASR model.
\end{abstract}
%

	\section{Introduction} \label{sec:intro}
	Speech Translation (ST) is the task of translating speech in a source language into text in a target language. Traditional ST approaches include two cascaded steps: Automatic Speech Recognition (ASR) and Machine Translation (MT). These approaches are prone to errors propagated from the ASR step to the MT step \cite{CascadedError}. Due to that, end-to-end ST has recently been gaining more interest \cite{ExampleDirect}, \cite{bansal-etal-2019-pre}, \cite{jia2019leveraging}, \cite{tang2021general}. End-to-end ST translates source-language speech directly into target-language text, thus less prone to error propagation. However, end-to-end ST performance heavily depends on direct ST data, which can be difficult to obtain. ASR and MT data, on the other hand, is often more easily available \cite{sperber-paulik-2020-speech}.

	To tackle data scarcity, it is useful to make use of ASR and MT data for end-to-end ST models. In order to achieve that, we explore techniques from zero-shot multilingual text translation and apply them to speech translation in this paper. Zero-shot, in the context of translation models, is an approach that enables translating a pair of languages that is unseen during training \cite{sperber-paulik-2020-speech}. We adapt this idea to speech translation by training a single multi-task model on ASR and MT data to perform the ST task in an end-to-end manner. We focus on a few-shot scenario, when a limited amount of ST data is available. In this scenario, we fine-tune our multi-task model with a small amount of ST data before performing ST task. 
	
	Motivated by the findings in multilingual text translation, we address two challenges to increase the efficiency of exploiting the ASR and MT data for ST task. First, we address the representation of semantically similar input using different modalities. Second, we address the model's ability to control the output language. We tackle the first challenge by using the Transformer architecture with shared encoder, and use an auxiliary loss function to minimize the difference in text and audio representation \cite{ZS-Quan}. For the second challenge, we propose a data augmentation approach to control the output language. We find these approaches to be particularly useful in the few-shot scenario. Our multi-task few-shot models outperform direct ST models trained on the same limited ST data from scratch by up to +12.9 BLEU points. Our models also outperform ST models fine-tuned from an ASR model using the same ST data by up to +3.1 BLEU points. This proves that our models have successfully made use of ASR and MT data to improve ST performance. The implementation is at \url{https://github.com/TuAnh23/MultiModalST}.

	\section{Related work} \label{sec:RelatedWork}
	Different approaches for ST have been explored over the decades, as summarized in \cite{sperber-paulik-2020-speech}. End-to-end ST, which is expected to overcome the error-propagation issue of the traditional cascaded approach, has recently been the method of interest. Examples of end-to-end ST include direct ST models trained on ST data from scratch \cite{ExampleDirect}, models pre-trained on ASR task and fine-tuned on ST task \cite{bansal-etal-2019-pre}, models trained on ST data generated by augmenting ASR or MT data \cite{jia2019leveraging} and models co-trained on MT and ST task \cite{tang2021general}. A major challenge of end-to-end ST is the lack of direct ST data for training.
	
	Zero-shot translation has been shown to work for multilingual  MT. In \cite{googleNMT}, zero-shot multilingual MT is enabled by adding a language token to the beginning of the input sequence to indicate the required target language. Several studies have been done to improve the quality of zero-shot multilingual MT. Approaches to encourage a source-language-independent representation are proposed in \cite{ZS-Quan}, such as using a fixed size encoder for different languages. In \cite{DEPI}, a language-independent representation is encouraged by disentangling positional information of the input and output tokens.
	
	Inspired by the ability of zero-shot for multilingual MT, we study the applicability of similar approaches on ST tasks, by building an ST multi-task model trained on ASR and MT tasks, which data is often more easily available than ST.
	
	\section{Multi-task model} \label{sec:zsST}
\begin{figure}[htb]
		\centerline{\includegraphics[width=0.26\textwidth]{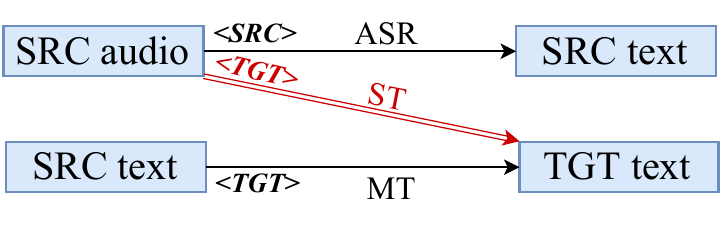}}
		\caption{Proposed multi-task model. Single-line arrows are training directions. Double-line arrow is inference direction. Tags in the brackets are target-language tokens.
		}
		\label{fig:baseZS}
	\end{figure}

	We propose a multi-task model as illustrated in Fig. \ref{fig:baseZS} to reduce the need for direct ST training data. The model is trained simultaneously on ASR task and MT task. We consider the few-shot scenario when a limited amount of ST data is available. We build few-shot model by fine-tuning the above multi-task models with a small amount of ST data before performing the ST task. We also attempt on zero-shot ST, using the multi-task model directly for ST task without fine-tuning. The requirement is that the model represents \textit{SRC audio} and \textit{SRC text} in a similar way so that it can leverage the ASR and MT tasks learnt during training to perform the ST task.

\begin{figure}[htb]
		\centerline{\includegraphics[width=0.49\textwidth]{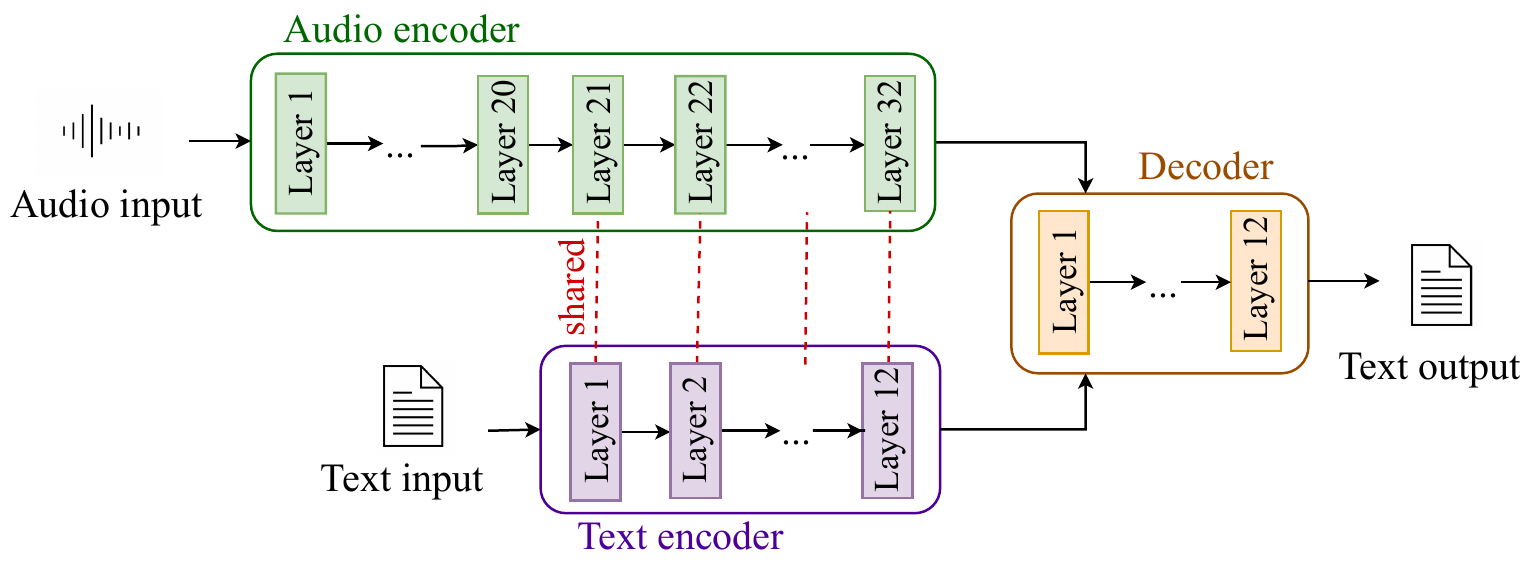}}
		\caption{Overall structure of multi-task models.}
		\label{fig:OverallStructure}
	\end{figure}

	We build upon the deep Transformer \cite{attention} for speech proposed in \cite{quan-2019-very-deep}. To enable encoding audio and text jointly, we share the model parameters between the two modalities. As we expect the bottom layers of the audio encoder to extract low-level acoustic features, we assign more layers to the audio encoder and share its top layers with the text encoder. This shared encoder is crucial to make audio and text representations similar. The overall structure is shown in Fig. \ref{fig:OverallStructure}. To indicate output language, we apply the same method as for zero-shot multilingual MT. We add target-language tokens to the beginning of input sequences and concatenate the target language embeddings to every decoder input to enforce the model outputting the language of interest \cite{ZS-Quan}. Target-language tokens are shown in the brackets in Fig. \ref{fig:baseZS}.
	
	The proposed model introduces a challenge, where it outputs the wrong language for ST task. It outputs \textit{SRC text} for \textit{SRC audio} input, even when the target token is \textit{$<$TGT$>$}. The reason is that in the training data, audio input always has \textit{SRC} output and text input always has \textit{TGT} output, thus the models decide on output language based on input modality instead of the target-language tokens. This boils down to two issues: the difference between modalities (text and audio), and the reliance of the model on the target-language tokens. We propose approaches to tackle these issues as follows.
	
	\subsection{Cross-modality knowledge sharing} \label{sec:auxloss}
\begin{figure}[htb]
		\centerline{\includegraphics[width=0.33\textwidth]{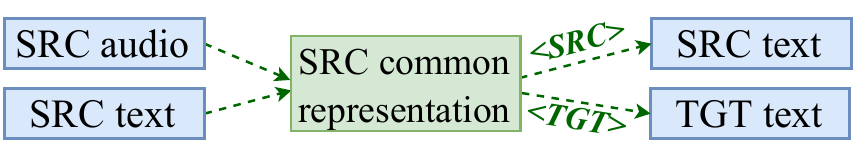}}
		\caption{Motivation for cross-modality knowledge sharing.}
		\label{fig:auxloss}
	\end{figure}

	A prerequisite to knowledge sharing across the ASR and MT tasks is common representations across input modalities, as shown in Fig. \ref{fig:auxloss}. We encourage modality-independent representation of the data by using an auxiliary loss function that minimizes the difference between encoder output of audio and text. The metric for the difference is the squared error of mean-pool over time \cite{ZS-Quan}. 
	That is, given a pair of aligned text sentence X and audio sentence Y, the auxiliary loss is:
	$[ mean\_pool(Encoder(X)) - mean\_pool(Encoder(Y)) ]^2
	$.

	\subsection{Controlling output language} \label{sec:data_aug}
	A challenge for the model in Fig. \ref{fig:baseZS} is to control the output language. As mentioned above, each modality has only one target language during training, leading to the model deciding on output language based on input modality, instead of the specified target-language tokens as desired.
	
	To have stronger control on the output language, we propose data augmentation to increase the model's reliance on the provided target tokens. The idea is to include artificial data so that each modality has more than one output language during training. We expect this to force the model to rely on the target tokens to decide which language to output. Data augmentation would avoid the need for another real dataset. 
	
\begin{figure}[htb]
		\centerline{\includegraphics[width=0.33\textwidth]{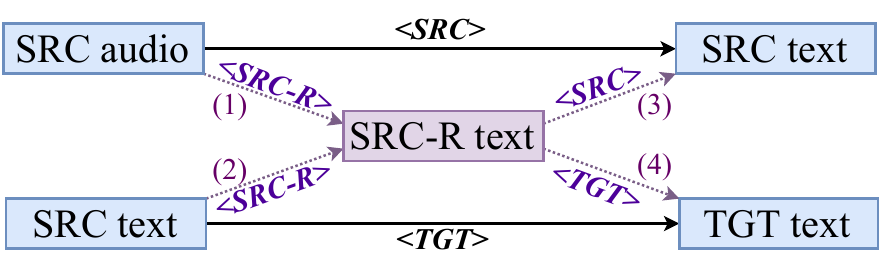}}
		\caption{Augmented training data. Solid-line arrows are main directions. Dotted-line arrows are artificial directions.}
		\label{fig:AD4}
	\end{figure}

    We introduce an artificial language by reversing \textit{SRC} sentences character-wise, denoted as \textit{SRC-R}. By reversing character-wise, the new language's vocabulary would not conflate with the original language. The training data is shown in Fig. \ref{fig:AD4}, along with the annotations. By adding (1) and (2), \textit{SRC audio} and \textit{SRC text} have two target languages during training. By adding (3) and (4), the model can learn to switch between outputting \textit{SRC} and \textit{TGT} languages. 
	
	\section{Experimental setup} \label{sec:exp_setup}
	The experiments goal is to build and evaluate ST models to translate English (\textit{EN}) audio to German (\textit{DE}) text. 
	
	\subsection{Dataset and preprocessing}
	We use the CoVoST 2 speech translation corpus \cite{wang2020covost}. The data used is English to German pair, with 289K training samples (430 hours of speech), 15K validation samples (26 hours of speech) and 15K testing samples (25 hours of speech).
	
	For audio data, we extract and normalize 40-dimensional log scale mel filterbank concatenated with its delta coefficient to use as input features. For text data, we remove the double quotes at the beginning and the end of the sentences, and use SentencePiece \cite{sentencepiece} without pre-tokenization and pre-normalization to build subword-based vocabularies. 
	
	\subsection{Model configurations}
	Our models use the Transformer architecture with attention-based encoder and decoder \cite{attention, quan-2019-very-deep}. 
	We adapt the hyperparameters choices in \cite{pham2019iwslt} to our multi-modal setting: 32 audio encoder layers, 12 text encoder layers, 12 decoder layers; the 12 text encoder layers are shared with the top 12 audio encoder layers. The other configurations are the same as \cite{pham2019iwslt}.

	When training multiple datasets in one model (e.g., ASR and MT), the batches from each dataset are ordered alternatively. The models are trained for 64 epochs, except for the fine-tuned models, where we stop training as soon as the validation loss stops reducing. The checkpointed model with the lowest validation loss is used for final evaluation.
	
	For ASR tasks, the models are evaluated using lowercased, tokenized, punctuation-removed Word Error Rate (WER) calculated using VizSeq \cite{wang2019vizseq}. For MT and ST tasks, the metric used is the detokenized, case-sensitive BLEU score, calculated using sacreBLEU \cite{bleu}. 
	
	\section{Results and Discussion} \label{sec:results}
	The transcription and translation quality of our few-shot models is shown in Table \ref{tab:overall}. In this few-shot scenario, we only use a limited amount (10\% and 25\%) of ST data. The amount of ASR and MT data used is 100\%.

	\begin{table*}[htb]
		\caption{Performance summary of different model types.}
		\begin{center}

            \begin{tabular}{llllll} 
            \toprule
            No. & Model type                                                                                                                                                                                       & \begin{tabular}[c]{@{}c@{}}ASR (WER $\downarrow$)\\(\%)\end{tabular} & \begin{tabular}[c]{@{}l@{}}MT (BLEU $\uparrow$)\\ $ $\end{tabular} & \begin{tabular}[c]{@{}l@{}}Few-shot ST (BLEU $\uparrow$) $^a$\\ 10\% ST data\end{tabular} & \begin{tabular}[c]{@{}l@{}}Few-shot ST (BLEU $\uparrow$) $^b$\\ 25\% ST data\end{tabular}  \\
            \midrule 
            1                                                              & \begin{tabular}[c]{@{}l@{}}Single ST task /\\Direct end-to-end ST\end{tabular}                                                                                                                   & -                                                                & -                                                              & ~~0.5                                                                         & ~~0.8 \\ 
    
            2                                                       & Single ASR task                                                                                                                                                                                  & 25.8                                                             & -                                                              & ~~8.4                                                                         & 10.9 \\ 
            3                                                        & Plain multi-task (P)                                                                                                                                                                             & 28.4                                                             & 32.8                                                           & ~~9.8                                                                         & 12.4 \\ 
    \midrule
            4                                                         & P + auxiliary loss                                                                                                                                                                               & 26.9                                                             & 32.5                                                           & 10.6 \textbf{(+10.1 $\vert$ +2.2 $\vert$ +0.8)}                                                                                          & 13.2 \textbf{(+12.4 $\vert$ +2.3 $\vert$ +0.8)} \\ 
            5                                                          & P + augmented data                                                                                                                                                                               & 47.6                                                             & 31.6                                                           & ~~5.4~ ~(\textbf{+4.9} $\vert$ ~-3.0 $\vert$ -4.4)                                                 & ~ 6.4~ ~(\textbf{+5.6} $\vert$ ~-4.5 $\vert$ -6.0) \\ 
            6                                                   & \begin{tabular}[c]{@{}l@{}}P + augmented data\\(fine-tune from plain)\end{tabular}                                                                       & 27.6                                                             & 32.2                                                           & 11.5~\textbf{\textbf{(+11.0 $\vert$ +3.1 $\vert$ +1.7)}}                                                                 & 13.5~\textbf{\textbf{(+12.7 $\vert$ +2.6 $\vert$ +1.1)}} \\ 
            7                                                           & \begin{tabular}[c]{@{}l@{}}P + augmented data~\\~ ~+ auxiliary loss\\(fine-tune from plain)\end{tabular} & 27.7                                                             & 32.3                                                           & 11.5~\textbf{\textbf{(+11.0 $\vert$ +3.1 $\vert$ +1.7)}}                                                                 & 13.7 \textbf{(+12.9 $\vert$ +2.8 $\vert$ +1.3)} \\ 
            \bottomrule
            \multicolumn{6}{l}{\begin{tabular}[c]{@{}l@{}}$^{a,b}$ Additionally reporting the differences compared to the three baselines in Row 1, Row 2 and Row 3, respectively.\end{tabular}}                                                                                                                                                                                     
            \end{tabular}
			\label{tab:overall}
		\end{center}
	\end{table*}
	
	We present three baselines using the same limited ST data: 
	(1) direct end-to-end ST model trained on ST data from scratch, (2) ST model fine-tuned from a pre-trained ASR model (indirectly making use of ASR data) and (3) ST model fine-tuned from plain multi-task model (indirectly making use of ASR and MT data).
	As shown in Row 1 of Table \ref{tab:overall}, the direct end-to-end ST model fails to perform the ST task given the extremely limited ST data. In contrast, the ST model making use of ASR data is able to perform ST task (Row 2). Our plain few-shot model, additionally making use of the MT data, has better ST performance, although the gain from the MT data is quite limited (Row 3).
	
	In Row 4-7, we denote our approaches with P + \{\textit{approach name}\}. We observe that our approaches improve the ST performance of the few-shot multi-task model. Data augmentation (Row 6) gives higher improvement than auxiliary loss (Row 4). The combination of both approaches (Row 7) gives the best performance, with the improvement of up to +12.9 BLEU points compared to the direct end-to-end baseline, +3.1 BLEU points compared to the fine-tuned ASR baseline and +1.7 compared to the plain multi-task baseline.
	We also observe that, as we increase the amount of ST data (25\% ST data instead of 10\%), the gains from our approaches generally become less significant. This suggests that our approaches are particularly effective in low-resource scenarios. 
	
	A challenge of data augmentation is that the model cannot learn ASR task well, leading to poor few-shot ST performance (Row 5), due to the high number of tasks in augmented data. Therefore, instead of training the model on augmented data from scratch, we first train it on the original data (ASR and MT), and then fine-tune with augmented data. In this way, the issue no longer persists (Row 6 and 7). It seems important for the model to first learn the easier task (ASR) and then ST task. This can be viewed as a type of curriculum learning.
	
	We also experimented on a direct ST model trained on 100\% of the ST data for comparison. The BLEU score of this model is 14.9. The BLEU score of our best few-shot model is 13.7. Our best few-shot model uses only 25\% of the ST data for training, yet only falls short by 1.2 BLEU points compared to this direct model using 100\% of the ST data.
	
	We also attempted on zero-shot using no ST data. This remains challenging, since the BLEU scores are under 1. However, the models using data augmentation no longer output all wrong language (as discussed in Section \ref{sec:zsST}), proving the effectiveness of data augmentation.

	\section{Analysis} \label{sec:analysis}
	\subsection{Cross-modal similarity and translation quality}
	We study the similarity of representations between modalities under different approaches and analyze them in relation to few-shot ST performance. We measure representational similarity between text and audio encoder output using Singular Vector Canonical Correlation Analysis (SVCCA) \cite{NIPS2017_7188}, where higher scores indicate higher similarity.
	
\begin{figure}[h]
		\centerline{\includegraphics[width=0.49\textwidth]{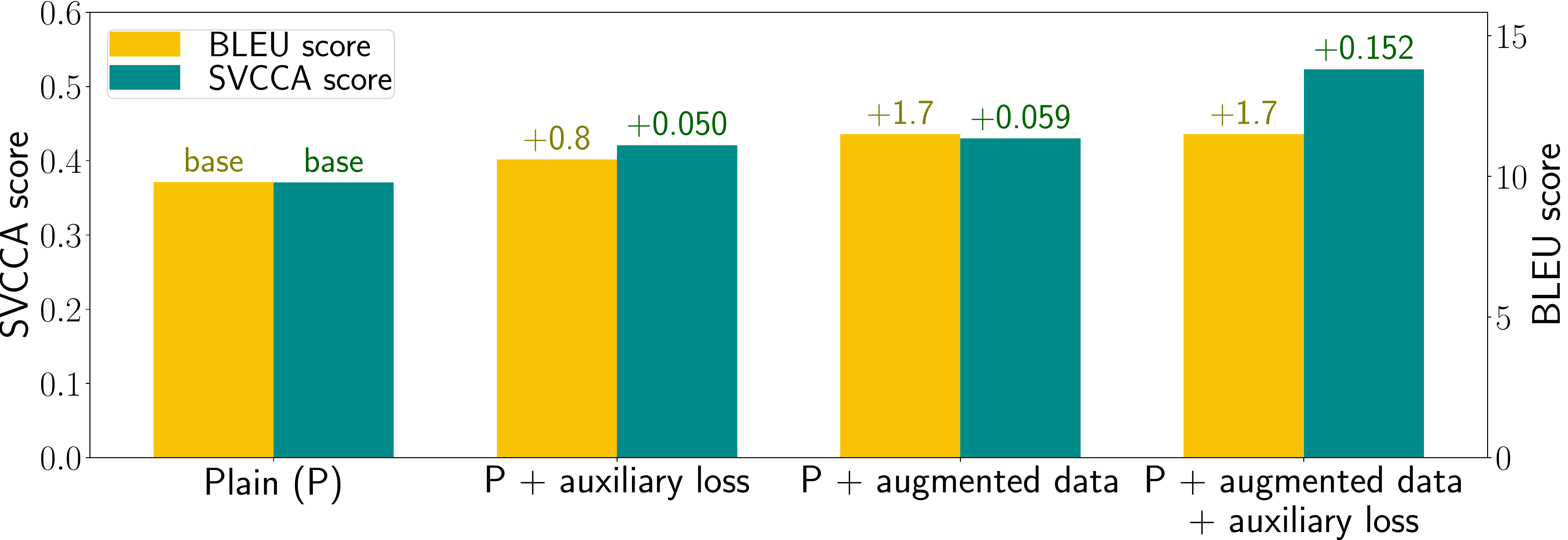}}
		\caption{SVCCA analysis along with translation quality. Numbers on the bars are comparision to the plain model.}
		\label{fig:modelsfull}
	\end{figure}

	The results are summarized in Fig. \ref{fig:modelsfull}, associating the similarity scores with few-shot translation quality. Observe that all proposed approaches increase text-audio similarity. The results also agree with our hypothesis: more text-audio similarity means higher BLEU score, i.e., better ST performance.

	\subsection{Cross-modal similarity at token and frame level}
	In addition to the SVCCA score, which measures text-audio similarity on a sentence level, we use another method to quantify text-audio similarity on a token level. We train a classifier to predict input modalities based on encoder outputs. Better classification performance indicates higher dissimilarity between the modalities \cite{DEPI}. Since the number of audio frames is significantly higher than the number of text tokens, we consider the True Positive Rate (TPR, proportion of audio tokens identified correctly) and the True Negative Rate (TNR, proportion of text tokens identified correctly) instead of the overall accuracy. The result is shown in Table \ref{tab:modality_classifiers}. We observe that for all models that do not use auxiliary loss, both the TPR and TNR are over 99.9\%, meaning that text and audio encoder output tokens are very distinguishable. On the other hand, for all models using auxiliary loss, the TPR is over 99.9\% and the TNR is under 10\%. This means that the classifier has a poor performance where it predicts most of the tokens as audio, which indicates high similarity between text and audio encoder output tokens. Thus, we conclude that using auxiliary loss indeed increases text-audio similarity on a token level.
	
	\begin{table}[htb]
		\caption{Performance of token modality classifier on the encoder outputs of different multi-task models.
		}
		\begin{center}
            \begin{tabular}{@{}lrr@{}}
            \toprule
                                                             Multi-task model                    & TPR (\%)    & TNR (\%)    \\ \midrule
            Plain multi-task (P)                                                                     & 99.99 & 99.89 \\
            P + augmented data                                                            & 99.99 & 99.99 \\
            P + auxiliary loss                                                            & 99.71 & \textbf{9.77} \\
            \begin{tabular}[c]{@{}l@{}}P + augmented data + auxiliary loss\end{tabular} & 99.99 & \textbf{.82} \\ \bottomrule
            \end{tabular}
			\label{tab:modality_classifiers}
		\end{center}
	\end{table}
	
	\section{Conclusions} \label{sec:conclusions}
	In this paper, we study how to alleviate the data scarcity problem of end-to-end ST by utilizing zero-shot multilingual MT techniques. Based on a multi-task model for ASR and MT, we propose approaches to (1) encourage knowledge sharing between text and audio modalities and (2) enforce stronger control of the output language. Our approaches successfully make use of ASR and MT data in the few-shot scenario, and improve the ST performance by up to +12.9 BLEU points compared to direct end-to-end ST models and +3.1 BLEU points compared to ST models fine-tuned from ASR models.

\bibliographystyle{IEEEbib}
\bibliography{references}

\end{document}